\documentclass[9pt,twocolumn,twoside]{opticajnl}
\journal{opticajournal}
\setboolean{shortarticle}{false} 

\usepackage{graphicx}
\usepackage{booktabs} 
\usepackage{hyperref} 

\title{Physics-consistent deep learning for blind aberration recovery in mobile optics}

\author[1,2]{Kartik Jhawar}
\author[1]{Tamo Sancho Miguel Tandoc}
\author[1]{Khoo Jun Xuan}
\author[1,2,*]{Wang Lipo}

\affil[1]{Institute for Digital Molecular Analytics and Science (IDMxS), Nanyang Technological University, Singapore 636921}
\affil[2]{School of Electrical and Electronic Engineering, Nanyang Technological University, Singapore 639798}

\affil[*]{elpwang@ntu.edu.sg}

\begin{abstract}
Mobile photography is often limited by complex, lens-specific optical aberrations. While recent deep learning methods approach this as an end-to-end deblurring task, these "black-box" models lack explicit optical modeling and can hallucinate details. Conversely, classical blind deconvolution remains highly unstable. To bridge this gap, we present \emph{Lens2Zernike}, a deep learning framework that blindly recovers physical optical parameters from a single blurred image. To the best of our knowledge, no prior work has simultaneously integrated supervision across three distinct optical domains. We introduce a novel physics-consistent strategy that explicitly minimizes errors via direct Zernike coefficient regression ($z$), differentiable physics constraints encompassing both wavefront and point spread function derivations ($p$), and auxiliary multi-task spatial map predictions ($m$). Through an ablation study on a ResNet-18 backbone, we demonstrate that our full multi-task framework ($z+p+m$) yields a 35\% improvement over coefficient-only baselines. Crucially, comparative analysis reveals that our approach outperforms two established deep learning methods from previous literature, achieving significantly lower regression errors. Ultimately, we demonstrate that these recovered physical parameters enable stable non-blind deconvolution, providing substantial in-domain improvement on the patented Institute for Digital Molecular Analytics and Science (IDMxS) Mobile Camera Lens Database for restoring diffraction-limited details from severely aberrated mobile captures.
\end{abstract}

\setboolean{displaycopyright}{false}

\begin{document}

\maketitle

\section{Introduction}

Mobile photography has evolved rapidly, yet the physical constraints of smartphone form factors necessitate the use of compact, plastic lens stacks. Unlike precision-ground glass optics, these molded plastic lenses suffer from complex, high-order aberrations that vary significantly not only across different phone models but also between units of the same model due to manufacturing tolerances \cite{Ref_MobileOptics, Ref_PlasticLensMfg}. These lens-specific aberrations introduce spatially variant blur that degrades image quality, particularly limiting the performance of downstream restoration tasks. While computational photography pipelines attempt to mitigate these artifacts, the unknown nature of the Point Spread Function (PSF) for any given device makes this an ill-posed inverse problem.

Classically, this restoration is framed as a blind deconvolution problem, where the algorithm must simultaneously estimate the latent sharp image and the blur kernel (PSF). Early variational methods attempted to constrain this solution space using natural image priors \cite{Ref_Levin2009, Ref_BlindDeconv}. However, these approaches often become unstable under strong blur or noise and fail to generalize when the scene content varies unpredictably. More recently, Deep Learning (DL) has shifted the paradigm toward "end-to-end" image deblurring, where a Convolutional Neural Network (CNN) maps blurred inputs directly to sharp outputs \cite{Ref_DeblurGAN, Ref_EndToEndDL}. While effective on their training datasets, these "black-box" models often lack explicit optical modeling; they learn to hallucinate high-frequency details rather than physically inverting the optical degradation. Consequently, they lack the physical reliability required for rigorous image restoration, even within a specific domain of lenses.

To address this lack of physical grounding, we propose a physics-consistent deep learning framework that predicts optical aberration parameters rather than estimating a pixel-wise PSF or restoring the image directly. Leveraging the patented IDMxS Mobile Camera Lens Database, we model the aberrations using Zernike polynomials at the pupil plane. Our hypothesis is that by regressing the Zernike coefficients ($Z_2$--$Z_{37}$), we can enforce a physically valid reconstruction of the wavefront. This lens-aware parameterization ensures that the recovered PSF satisfies Fourier optics constraints, enabling stable and explainable downstream deconvolution.

In this work, we demonstrate a lens-indexed aberration recovery method that ensures robust in-domain generalization. Our core contribution is a physics-consistent supervision strategy that combines three complementary loss terms: (i) a Zernike regression loss in the normalized coefficient space ($z$); (ii) a wavefront and PSF reconstruction loss derived via a differentiable optics layer ($p$); and (iii) an auxiliary multi-head supervision that explicitly predicts wavefront maps ($m$). We validate this approach on in-domain lens designs that were strictly held out from the training set (i.e., unseen lenses from the same patented IDMxS Mobile Camera Lens Database). By doing so, we provide a rigorous architectural analysis of how these constraints interact with modern models to offer a robust alternative to purely data-driven deblurring.

\section{Materials and Methods}

\subsection{Dataset Generation}
To ensure physical realism, we utilized the patented IDMxS Mobile Camera Lens Database\footnote{\url{https://idmxs.org/research/mobile-phone-camera-lens-database/}} \cite{Ref_IDMxS_Database}, utilizing 109 discrete smartphone lens designs provided as Zemax OpticStudio files. We extracted the Zernike coefficients ($Z_2$--$Z_{37}$, following the standard Noll sequential indexing scheme) for each lens to characterize its specific aberration signature. Ground truth clean images were sourced from a public smartphone microscopy dataset \cite{Ref_MicroscopyData} to simulate high-frequency biological structures. We generated a training corpus of 110,090 synthetic blurred images by convolving clean patches ($256 \times 256$ pixels) with Point Spread Functions (PSFs) computed from the lens coefficients via a Fourier optics model. By explicitly modeling these complex, asymmetrical PSFs, this rigorous forward simulation ensures the network learns to invert true physical diffraction and lens-specific aberrations, rather than simply learning to sharpen the mathematically uniform, generic Gaussian blur commonly used in standard deep learning datasets.

\subsection{Physics-Consistent Network Architecture}
Our aberration recovery model, \emph{Lens2Zernike}, employs a ResNet-18 backbone modified to regress the 36-dimensional Zernike vector. To overcome the limitations of "black-box" regression, we introduce a physics-consistent supervision strategy combining three loss terms:
\begin{equation}
\mathcal{L}_{total} = \lambda_z \mathcal{L}_{coeff} + \lambda_p \mathcal{L}_{physics} + \lambda_m \mathcal{L}_{map}
\label{eq:losses}
\end{equation}
(i) \textbf{Coefficient Loss ($\mathcal{L}_{coeff}$):} A standard Mean Squared Error (MSE) loss in the normalized Zernike coefficient space.
(ii) \textbf{Physics Loss ($\mathcal{L}_{physics}$):} We implement a differentiable optics layer that maps predicted coefficients to a wavefront phase map ($\phi$) and subsequently to a PSF via the Fourier transform. We minimize the MSE between these derived physical quantities and their ground truths. This ensures that even if individual coefficients drift, the resulting optical effect remains constrained.
(iii) \textbf{Multi-task Map Loss ($\mathcal{L}_{map}$):} Auxiliary decoder heads explicitly predict the high-resolution wavefront and PSF maps, providing dense spatial supervision that guides the convolutional encoder during training.

While the network is optimized using MSE to penalize large outliers during training, our primary quantitative evaluation metrics are the Mean Absolute Error (MAE) and MSE computed in the unnormalized physical wave space ($\lambda$). MAE provides a direct, interpretable measure of average physical aberration severity. Additionally, we utilize Peak Signal-to-Noise Ratio (PSNR) to evaluate the downstream image restoration quality.

\section{Results}

We evaluated our model using 5-fold cross-validation. To ensure robust in-domain generalization, we strictly separated the lens designs between the training and testing sets, ensuring the network is evaluated on unseen lenses drawn exclusively from the same patented IDMxS Mobile Camera Lens Database. Throughout our evaluation, errors are reported in units of $\lambda$, representing the reference optical wavelength.

\subsection{Ablation Study: Physics-Consistent Supervision}
To isolate the contribution of our proposed supervision strategy, we performed a granular ablation study on the ResNet-18 backbone. We sequentially added the Wavefront loss ($p_W$), the PSF loss ($p_P$), combined them ($p$), and finally added the multi-task heads ($m$).

The results in Table \ref{tab:ablation} confirm that grounding predictions in the physical domain is vastly superior to purely coefficient-based regression. While adding Wavefront ($p_W$) or PSF ($p_P$) constraints individually improved performance, combining them ($p$) yielded a stronger reduction in Mean Absolute Error (MAE). The best performance was achieved by the full multi-task model ($z+p+m$), which provided the final refinement to $0.00128\lambda$, representing an approximate 35\% improvement over the baseline.

\begin{table}[htbp]
\centering
\caption{\bf Component Ablation Study (ResNet-18)}
\begin{tabular}{lcc}
\toprule
Model Variant & MAE ($\lambda$) & MSE ($\lambda^2$) \\
\midrule
$z$ (Baseline Coeff-only) & $0.00197$ & $5.82 \times 10^{-5}$ \\
$z + \mathcal{L}_{wave}$ ($p_W$ only) & $0.00156$ & $5.56 \times 10^{-5}$ \\
$z + \mathcal{L}_{psf}$ ($p_P$ only) & $0.00152$ & $5.19 \times 10^{-5}$ \\
$z + p$ (Combined Physics) & $0.00137$ & $4.73 \times 10^{-5}$ \\
\textbf{$z+p+m$ (Full Multi-task)} & \textbf{0.00128} & \textbf{4.20 $\times 10^{-5}$} \\
\bottomrule
\end{tabular}
\label{tab:ablation}
\end{table}

\begin{figure}[ht!]
\centering
\includegraphics[width=0.9\linewidth]{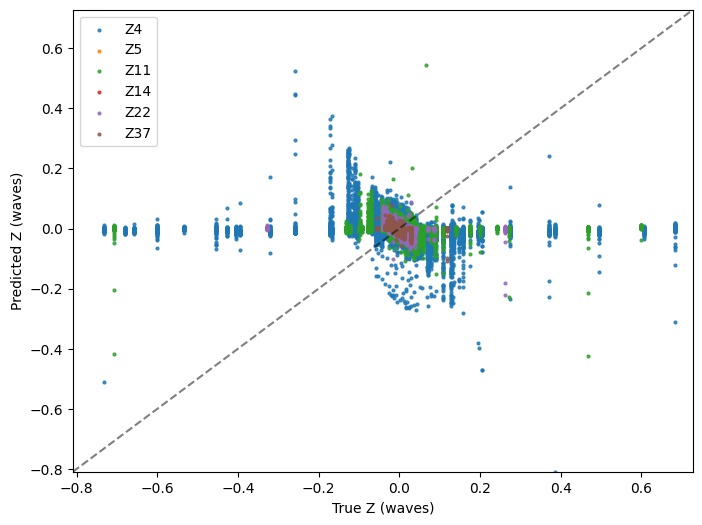}
\caption{Prediction accuracy. Scatter plot of predicted vs. true Zernike coefficients for the test set. The tight correlation confirms the model effectively learns diverse aberration modes.}
\label{fig:accuracy}
\end{figure}

\subsection{Comparative Analysis}
To benchmark our approach against existing methods, we adapted DLWFS (based on Xception) \cite{nishizaki2019deep} and DLAO (LAPANet) \cite{song2025dlao}. Table \ref{tab:sota} demonstrates that our physics-aware ResNet-18 ($m$) demonstrates highly competitive performance, achieving a distinctly lower regression error (MAE $0.00128\lambda$) compared to the adapted DLWFS (MAE $0.00173\lambda$) and decoding-based DLAO (MAE $0.00324\lambda$) baselines.

\begin{table}[htbp]
\centering
\caption{\bf Comparison with Literature Baselines}
\begin{tabular}{lcc}
\toprule
Method & Backbone & MAE ($\lambda$) \\
\midrule
DLAO (2025) \cite{song2025dlao} & LAPANet & $0.00324$ \\
DLWFS (2019) \cite{nishizaki2019deep} & Xception & $0.00173$ \\
\textbf{Ours ($z+p+m$)} & \textbf{ResNet-18} & \textbf{0.00128} \\
\bottomrule
\end{tabular}
\label{tab:sota}
\end{table}

\subsection{Downstream Image Restoration}
\begin{figure}[ht!]
\centering
\includegraphics[width=\linewidth]{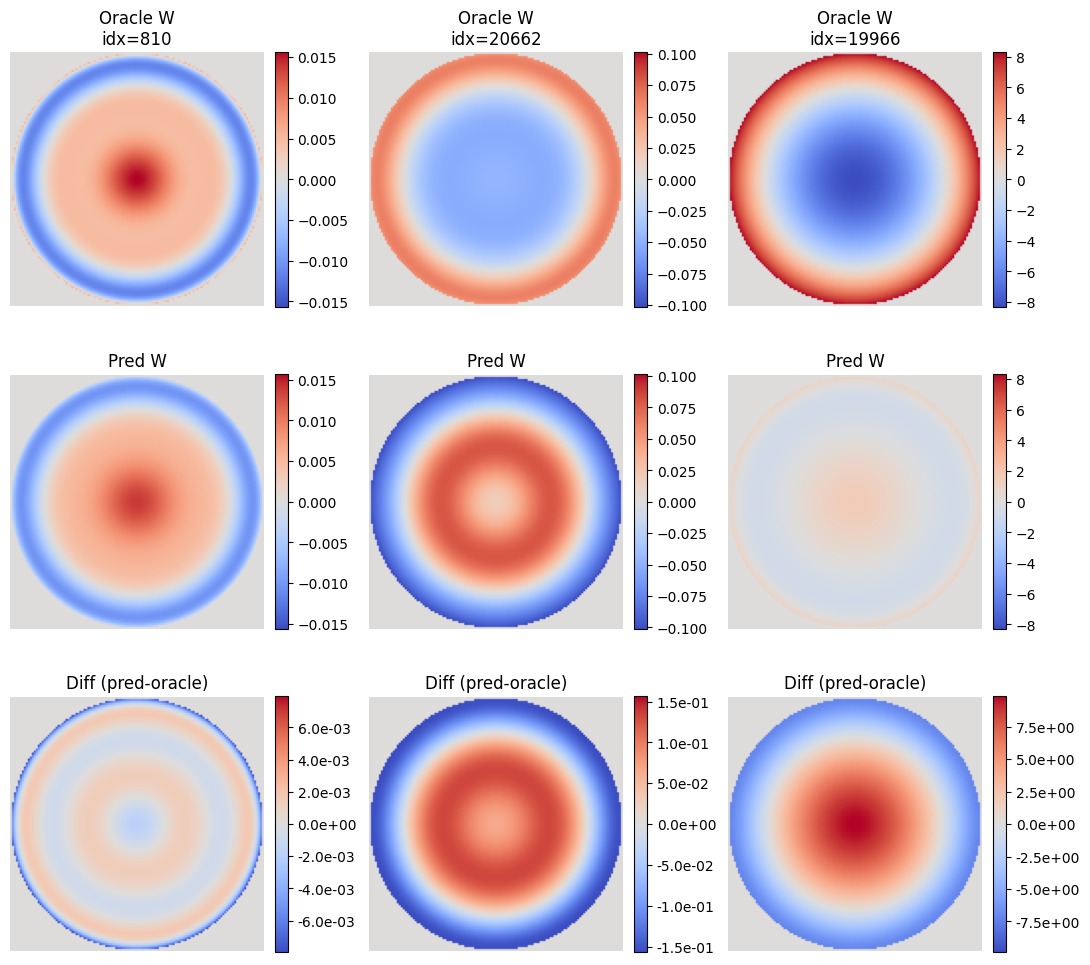}
\caption{Wavefront fidelity. Visual comparison of Oracle (top) vs. Predicted (middle) wavefront maps for three test cases (columns). The difference maps (bottom) visually highlight structural residuals; quantitatively, the absolute magnitude of these residuals is highly constrained, indicating accurate physical reconstruction.}
\label{fig:wavefront}
\end{figure}

\begin{figure}[ht!]
\centering
\includegraphics[width=\linewidth]{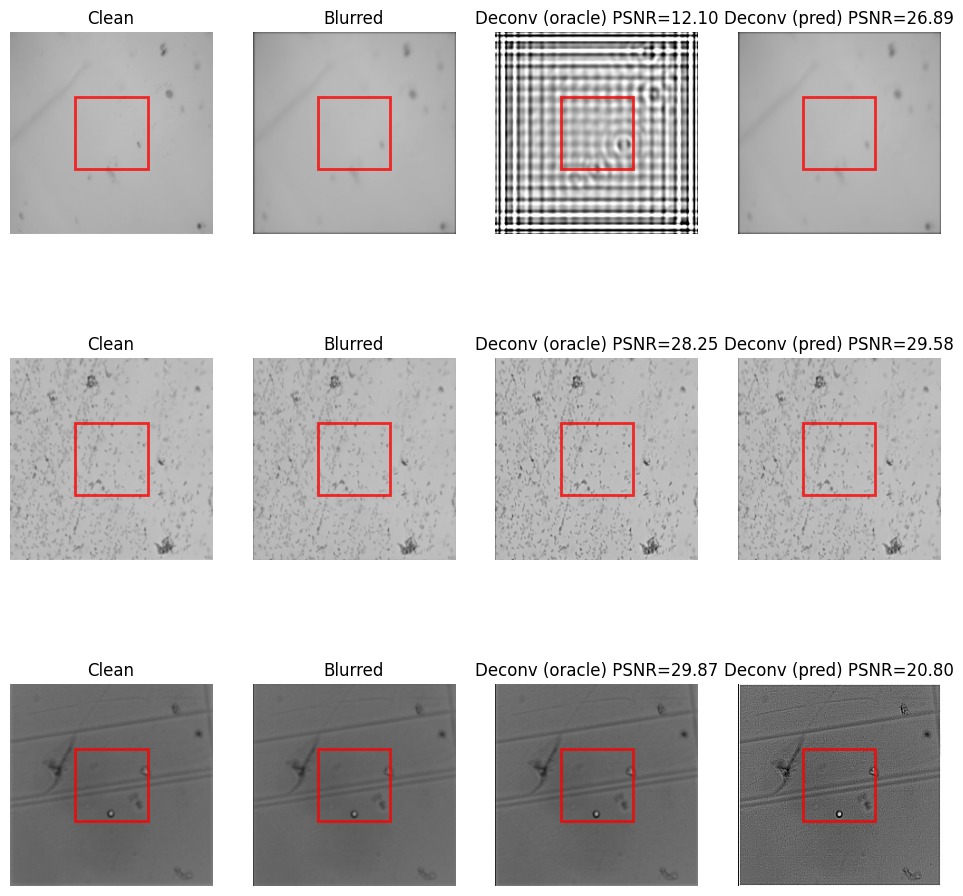}
\caption{Restoration performance. Qualitative comparison of deconvolution results. The predicted PSFs successfully recover fine structures and cellular boundaries obscured by lens blur (column 2), achieving clarity and structural fidelity comparable to the Oracle deconvolution (column 3). Highlights indicate regions of significant high-frequency recovery.}
\label{fig:restoration}
\end{figure}

The practical utility of our method is evident in restoration tasks. Figure \ref{fig:wavefront} visualizes the fidelity of the reconstructed wavefronts. Furthermore, as shown in Figure \ref{fig:restoration}, we performed non-blind Wiener deconvolution using the PSFs derived from our predictions. The restoration achieved a mean PSNR of 24.66 dB on the test set, closely matching the Oracle (Ground Truth PSF) restoration of 25.02 dB. The narrow "Oracle Gap" ($-0.36$ dB) indicates that the predicted Zernike vector captures the primary aberrations governing image degradation.

\section{Discussion}

The results from our ablation study highlight the critical importance of domain-specific optical constraints when regressing complex aberration parameters. 

The pure physics loss ($p$) acts as an effective physical regularizer. Because it mathematically computes the PSF and wavefront directly from the 1D Zernike vector without requiring additional network parameters, it grounds the feature extraction process in realistic optical behavior, significantly reducing regression error compared to purely data-driven coefficient mapping ($z$).

Furthermore, introducing the auxiliary multi-task maps ($m$) forces the network to explicitly reconstruct dense 2D spatial features alongside the 1D global coefficients. Standard CNNs, like our chosen ResNet-18 backbone, possess a continuous spatial structure that harmonizes well with this dense decoding task. 

Therefore, we find that while the $z+p$ supervision provides a strong, physics-grounded baseline for optical recovery, pairing the full $z+p+m$ constraint with the spatial bias of a traditional CNN offers a highly effective parameterization of complex lens aberrations, outperforming established methods in the literature.

\section{Conclusion}

We have demonstrated a physics-consistent deep learning approach for blind aberration recovery in mobile optics. By utilizing the patented IDMxS Mobile Camera Lens Database to train on physically rigorous data, and by enforcing consistency across the coefficient, wavefront, and PSF domains, we achieved a highly robust solution that generalizes effectively to in-domain, held-out lens designs. 

Unlike "black-box" image restoration methods, our \emph{Lens2Zernike} pipeline produces an explainable optical parameterization (Zernike coefficients), allowing for flexible downstream applications ranging from deconvolution to digital aberration correction. Future work will explore validating this pipeline on captured hardware data and expanding the Zernike order to model increasingly complex plastic lens deformations.

\begin{backmatter}
\bmsection{Funding}
This research is supported by the Ministry of Education, Singapore, under its Research Centre of Excellence award to the Institute for Digital Molecular Analytics \& Science, NTU (IDMxS, grant: EDUNC-33-18-279-V12)

\bmsection{Acknowledgment}
The authors thank the IDMxS technical team for providing access to the IDMxS Mobile Camera Lens Database and computing resources. T.S.M. Tandoc and Jun Xuan acknowledges the support of the Industry Immersion Programme (IIP).

\bmsection{Disclosures}
The authors declare no conflicts of interest.

\bmsection{Data Availability Statement}
Data underlying the results presented in this paper are based on the proprietary IDMxS Mobile Camera Lens Database, available at \href{https://idmxs.org/research/mobile-phone-camera-lens-database/}{https://idmxs.org/research/mobile-phone-camera-lens-database/}.
\end{backmatter}

\bibliography{sample}
\bibliographyfullrefs{sample}

\end{document}